\documentclass[sigconf]{acmart}
\AtBeginDocument{%
  }

\copyrightyear{2026}
\acmYear{2026}
\setcopyright{cc}
\setcctype{by}
\acmConference[ICTIR '26]{Proceedings of the 2026 International ACM SIGIR Conference on Innovative Concepts and Theories in Information Retrieval (ICTIR)}{July 25, 2026}{Melbourne, VIC, Australia}
\acmBooktitle{Proceedings of the 2026 International ACM SIGIR Conference on Innovative Concepts and Theories in Information Retrieval (ICTIR) (ICTIR '26), July 25, 2026, Melbourne, VIC, Australia}
\acmDOI{10.1145/3805713.3820439}
\acmISBN{979-8-4007-2600-2/2026/07}
  
\usepackage{multirow}
\usepackage{todonotes}
\usepackage[table]{xcolor}

\begin{document}

\title{Patent Representation Learning via Self-supervision}
\titlenote{This is a corrected and revised version of the paper published at ICTIR~'26 (DOI: 10.1145/3805713.3820439). It corrects a subset of dropout-only baseline results ($p$=0.2 and 0.3) affected by an implementation error and revises the methodology description (Sections~4--5.3) to match the released implementation; all conclusions are unchanged.}

%

\author{You Zuo}
\orcid{0000-0003-3951-7915}
\affiliation{%
  \institution{Inria}
  \city{Paris}
  \country{France}}
\affiliation{%
  \institution{Questel}
  \city{Paris}
  \country{France}}
\email{you.zuo@inria.fr}

\author{Kim Gerdes}
\orcid{0000-0002-9905-0117}
\affiliation{%
  \department{Laboratoire Interdisciplinaire des Sciences du Numérique (LISN), CNRS}
  \institution{Université Paris-Saclay}
  \city{Orsay}
  \country{France}}
\affiliation{%
  \institution{Questel}
  \city{Paris}
  \country{France}}
\email{gerdes@lisn.fr}

\author{\'{E}ric de la Clergerie}
\orcid{0000-0001-6428-9219}
\affiliation{%
  \institution{Inria}
  \city{Paris}
  \country{France}}
\email{eric.de_la_clergerie@inria.fr}

\author{Beno\^{\i}t Sagot}
\orcid{0000-0002-0107-8526}
\affiliation{%
  \institution{Inria}
  \city{Paris}
  \country{France}}
\email{benoit.sagot@inria.fr}


\begin{abstract}
We study self-supervised patent representation learning with contrastive objectives. A standard baseline constructs positives by encoding the same text twice under independent dropout masks, but applying this recipe to long, structured patent documents requires careful calibration. We show that dropout-only training can be substantially strengthened by tuning temperature and dropout rate, yet its best configuration is evaluation-dependent and does not transfer uniformly from title--abstract retrieval to claim-to-disclosure retrieval. We propose mixed dropout--section positives, a patent-specific view construction strategy in which each positive pair links the title--abstract view of a patent either to a dropout re-encoding of itself or to another section of the same patent, such as claims, summary, background, drawings, or description. This uses patent-internal structure as a training-time signal without IPC labels, citations, or relevance annotations. We evaluate on graded EPO search-report retrieval, DAPFAM, a recently proposed family-level patent retrieval benchmark, and IPC subclass classification. Section-based positives improve over calibrated dropout-only and generic title--abstract augmentation baselines, are competitive with citation-informed patent encoders and a general-purpose embedding model, and perform strongly on the out-of-domain split of DAPFAM. Additional cross-section alignment diagnostics show that section-pair training improves compatibility among abstracts, claims, and descriptions of the same invention. These results indicate that patent sections provide effective self-supervised positive views for learning dense patent representations.
\end{abstract}




\maketitle

\section{Introduction}

Patents constitute one of the largest structured repositories of technical knowledge: in 2024, innovators filed 3.7 million patent applications worldwide, bringing the total number of patents in force to about 19.7 million~\cite{wipo2025facts}. Their scale and legal--technical complexity make manual analysis increasingly difficult, creating a strong need for automated systems that can retrieve relevant prior art, organize technologies, and support downstream analytics. A central requirement for such systems is to learn patent representations that capture fine-grained semantic relatedness for prior-art search, while also preserving broader technology-domain structure for clustering and exploration.

Existing patent embedding methods often rely on external supervision, such as IPC\footnote{International Patent Classification: \url{https://www.wipo.int/en/web/classification-ipc}} labels~\cite{li2022copate} or citation graphs~\cite{vowinckel2023searchformer,ghosh2024paecter}. These signals are useful, but they are not the same as textual semantic relatedness. IPC codes provide coarse, manually assigned technology categories and may group broad technical areas rather than fine-grained relevance. Patent citations, in turn, may reflect institutional and procedural factors in search and examination, in addition to technological relatedness~\cite{criscuolo2008does}. This motivates self-supervised learning approaches that learn directly from patent text without IPC labels, citations, or external relevance annotations.

A natural self-supervised baseline for text embeddings is contrastive learning with dropout-based positives, following SimCSE~\cite{gao2021simcse}: the same input is encoded twice under independent dropout masks and treated as a positive pair, while other examples in the mini-batch serve as negatives. Applying this recipe to patents requires calibration. Patent inputs are often long and structured, and long-document embeddings can exhibit different similarity behavior from sentence embeddings~\cite{zhou2025length}. In our experiments, dropout-only training is sensitive to both the dropout rate and the temperature, and stronger calibration substantially improves some retrieval and classification settings. We therefore study dropout-only training as a strong tunable baseline rather than using a single default configuration as the point of comparison. This raises a complementary question: beyond tuning stochastic views of the same title--abstract input, can patent-internal structure provide positive views that remain aligned at the invention level while adding more stable document-level variation?

Patent-internal sections provide a natural source of such variation. Different sections of the same patent, such as the abstract, claims, summary, background, drawings, and description, describe the same invention from different legal and technical perspectives. They are aligned at the invention level, but differ in discourse role, length, and vocabulary. This structure is specific to patents and is available without external labels. We exploit it during training by constructing mixed dropout--section positives: each positive pair links the title--abstract view of a patent either to another dropout encoding of the same title--abstract text or to another section of the same patent, with the two branches sampled in equal proportion. The dropout branch preserves local stability under stochastic encoder noise, while the section branch exposes the model to discourse-level variation within the same invention, on top of the encoder-level stochasticity shared by all views.

We evaluate this strategy on prior-art retrieval and IPC subclass classification. For retrieval, we use both our EPO search-report benchmark and DAPFAM~\cite{ayaou2025dapfam}, a family-level patent retrieval benchmark with in-domain and out-of-domain settings. Our EPO benchmark uses examiner relevance categories and carefully designed hard negatives, allowing us to evaluate rank-aware retrieval with graded relevance. DAPFAM provides a complementary evaluation at the patent-family level with citation-based relevance judgments. We compare mixed dropout--section positives against lexical retrieval, patent-domain encoders, citation-informed models, a large general-purpose embedding model, calibrated dropout-only baselines, and generic text augmentations.

The results show that positive-view construction interacts with contrastive temperature and dropout rate. Dropout-only training becomes substantially stronger after calibration, especially for title--abstract retrieval and IPC classification, but no single dropout-only configuration performs best across all evaluation settings. Patent-section positives improve over dropout-only baselines and are competitive with strong retrieval baselines on EPO retrieval, while also performing well on DAPFAM family-level retrieval. The most effective sections differ by evaluation target: claim-based positives are most effective for prior-art retrieval, while background-oriented combinations better capture IPC subclass structure.

Overall, this work studies how to construct positive views for self-supervised patent representation learning. We revisit dropout-only contrastive learning under stronger calibration, introduce mixed dropout--section positives as a training-time use of patent structure, and evaluate the resulting representations on graded EPO search-report retrieval, DAPFAM family-level retrieval, and IPC KNN classification. Across these evaluations, patent-internal sections provide a useful source of positive-view variation, complementing dropout-based local consistency without requiring IPC labels, citations, or external relevance annotations during training. We release code, checkpoints, and evaluation queries for reproducibility: \url{https://github.com/ZoeYou/patentmapv0}.

\section{Related Work}

\paragraph{Patent representation learning.}
Patent representation learning has mainly relied on domain adaptation or external supervision. PatentBERT~\cite{lee2020patent} fine-tunes BERT on patent claims for IPC/CPC classification, while PatentSBERTa~\cite{bekamiri2024patentsberta} uses cross-encoder silver labels for patent similarity. BERT-for-Patents~\cite{SrebrovicYonamine2020} and Patent ModernBERT~\cite{yousefiramandi2025patent} pretrain encoders on large patent corpora to better model technical language. Other work uses patent metadata as training signals: CoPatE~\cite{li2022copate} constructs IPC-based pairs, while SEARCHFORMER~\cite{vowinckel2023searchformer}, Pat-SPECTER,  PaECTER~\cite{ghosh2024paecter}, and QaECTER~\cite{djemmal2026citationdrivenmultiviewtrainingpatent} use citation-informed supervision. Together, these models provide strong points of comparison: domain-pretrained encoders test the value of patent-specific language modeling, while supervised or weakly supervised methods test the value of IPC labels, citations, graphs, silver labels, or task-specific objectives. Our work instead studies whether patent-internal document structure can provide self-supervised training signals for dense patent embeddings, without IPC labels, citations, or external relevance annotations.

\paragraph{Structured patent retrieval.}
Patent retrieval has long exploited the internal structure of patent documents. Earlier work on invalidity and prior-art search decomposed long patent queries into subtopics or segments and combined the resulting retrieval streams~\cite{takaki2004associative,ganguly2011united}, or reduced verbose patent queries using pseudo-relevance feedback~\cite{ganguly2011patent}. Claim-to-passage matching also reflects the structure of examiner search reports: PatentMatch~\cite{risch2020patentmatch} operationalizes this setting as a supervised dataset of application claims paired with examiner-identified passages from cited documents. These studies show that patent structure matters for retrieval. Our use of structure is different: we do not use sections as retrieval-time fields, query reductions, or fusion units. This shifts the role of structure from retrieval-time decomposition and fusion to training-time view construction.

\paragraph{Self-supervised text embeddings.}
Contrastive representation learning constructs multiple views of the same input and trains an encoder to bring positives closer than negatives~\cite{chen2020simple}. In NLP, SimCSE~\cite{gao2021simcse} showed that independent dropout masks provide a simple positive-pair construction for sentence embeddings, while other work uses textual perturbations~\cite{wu2020clear,yan2021consert}, hard-negative strategies~\cite{wu2021smoothed,zhou2022debiased,deng2023clustering}, or modified contrastive objectives~\cite{fang2020cert,zhang2021bootstrapped,xu2023simcse++}. Recent long-document work points to additional issues when applying sentence-level embedding recipes to longer texts: dense long-document embeddings may need objectives or distances beyond cosine-based contrastive learning~\cite{saggau2023efficient}, and longer inputs can make PLM embeddings less discriminative through length-induced representation collapse~\cite{zhou2025length}. These findings suggest that positive-view construction for patents should account for both document length and internal structure. Our work studies this question in the patent domain by using patent sections as training-time positive views, rather than as retrieval-time fields or externally supervised labels.

\section{From Dropout Positives to Patent Sections}
\label{sec:section_views}

A central design choice in contrastive self-supervision is how to construct positive views. For text embeddings, a common strategy is to encode the same input twice under independent dropout masks, as in SimCSE~\cite{gao2021simcse}. This is also a natural baseline for patents: two encodings of the same title--abstract view refer to the same invention, while dropout introduces stochastic encoder-level variation. However, dropout positives keep the text fixed. For long and structured patent inputs, this can provide only limited view-level variation, so the two positive views may already have high similarity early in training. Under a default InfoNCE calibration, this can make the positive-pair signal less informative. Lowering the contrastive temperature partly changes this behavior by increasing the sharpness of the contrastive distribution, which makes dropout-only training a stronger baseline in our experiments.

\begin{figure}[ht]
\centering
\includegraphics[width=\linewidth]{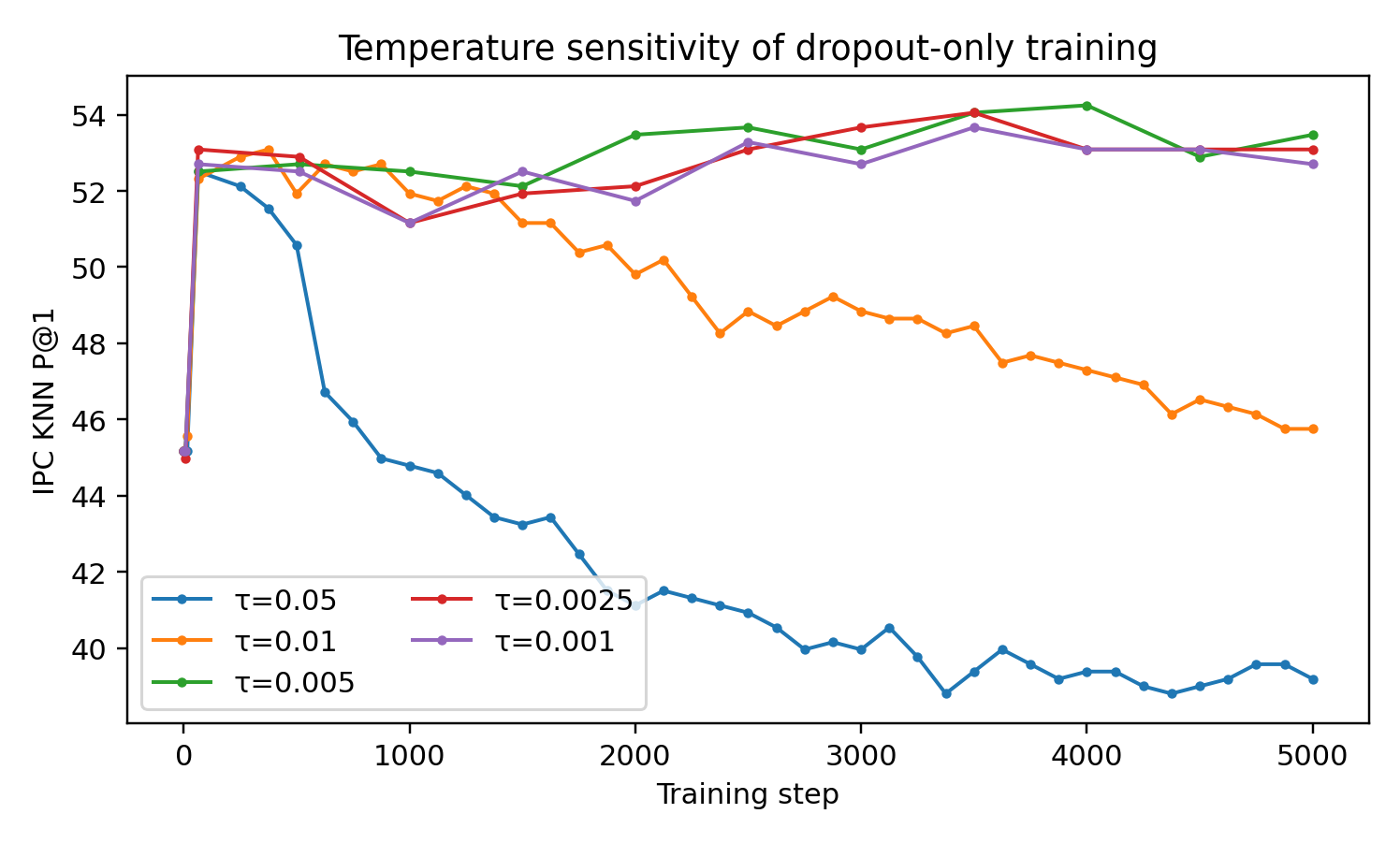}
\caption{
Temperature sensitivity of dropout-only contrastive training. 
We report IPC KNN P@1 during training with a fixed dropout rate of 0.1 and different contrastive temperatures. 
Dropout-only training is highly sensitive to temperature: the default setting peaks early and then declines, while lower temperatures yield stronger final checkpoints. 
}
\label{fig:dropout_temperature}
\end{figure}

Figure~\ref{fig:dropout_temperature} illustrates how temperature tuning affects dropout-only training. With the same dropout rate, changing only the contrastive temperature leads to different IPC KNN trajectories during early training. The default temperature improves early and then degrades, while lower temperatures lead to stronger checkpoints in this diagnostic. This does not imply that dropout-only training is ineffective; rather, it shows that dropout-only positives should be treated as a calibrated baseline instead of a single default configuration.
We revisit this calibration issue in Section~\ref{sec:temperature}.

Patent sections provide a complementary source of positive views. Patents describe the same invention through standardized sections with different discourse roles: the title and abstract summarize the invention, claims define its legal scope, and sections such as summary, background, drawings, and description provide technical context and implementation details. These sections are not interchangeable, but they are linked by the same underlying invention. They therefore offer positive views that preserve invention-level alignment while introducing variation in wording, length, and discourse role.

We use two principles to guide positive-view construction. First, positive views should preserve \emph{invention-level alignment}: both views should refer to the same invention rather than to merely related technologies. Second, they should provide \emph{view-level variation}: the views should differ enough in wording, length, or discourse role to make the training signal non-trivial. Dropout positives mainly provide stochastic encoder variation while keeping the text fixed. Cross-section positives preserve the same invention-level alignment, while adding discourse-level variation that is native to patent documents, on top of the encoder-level stochasticity that all views share during training.

Our method combines these two sources of variation. Dropout positives encourage local stability under stochastic encoder noise, while section positives expose the encoder to different textual realizations of the same invention. The goal is not to replace dropout with sections, but to add patent-specific positive views to a calibrated dropout baseline.

\section{Methodology}
\label{sec:method}

We propose a mixed dropout--section contrastive objective for self-supervised patent representation learning. The method uses only the text available within each patent document and does not require IPC labels, citation links, or relevance annotations during training.

\subsection{Patent Views}
For each patent $i$, we use the title and abstract as the default patent view in contrastive training:
\begin{equation}
x_i^{\mathrm{TA}} = [x_i^{\mathrm{title}}; x_i^{\mathrm{abstract}}].
\end{equation}
The title--abstract view provides a compact description of the invention and is available for most patent documents, making it a stable reference view for pairing with other patent sections during self-supervised training.

Let \(\mathcal{S}^{\mathrm{all}}\) denote the inventory of \emph{cross-section} views that is potentially available during training: claims, summary, background, detailed description, and drawing description. The title--abstract view itself is not part of this inventory; it enters the objective through a separate dropout branch described below. Each training variant specifies a subset \(\mathcal{S}^{(v)} \subseteq \mathcal{S}^{\mathrm{all}}\) from which cross-section views may be sampled. For patent $i$, the available section views under variant $v$ are
\begin{equation}
\mathcal{S}_i^{(v)} = \mathcal{S}^{(v)} \cap \mathcal{A}_i ,
\end{equation}
where \(\mathcal{A}_i\) is the set of sections present in patent $i$. Patents without the required title--abstract view are excluded from training, while missing optional sections are removed from \(\mathcal{S}_i^{(v)}\).

\subsection{Mixed Positive Construction}
\label{sec:mixed_construction}

Positive pairs are constructed by a two-level sampling procedure. For each training instance, the first level samples uniformly between the dropout strategy and the section strategy:
\begin{equation}
b_i \sim \mathrm{Unif}(\{\mathrm{dropout},\ \mathrm{section}\}),
\label{eq:strategy_sampling}
\end{equation}
and the section strategy then draws a concrete section type uniformly at a second level.

For the main method, the two views of patent $i$ are instantiated as
\begin{equation}
\left(x_i^{(1)},\, x_i^{(2)}\right) =
\begin{cases}
\left(x_i^{\mathrm{TA}},\ x_i^{\mathrm{TA}}\right), & \text{if } b_i = \mathrm{dropout},\\[2pt]
\left(x_i^{\mathrm{TA}},\ x_i^{s_i}\right),\quad s_i \sim \mathrm{Unif}\!\left(\mathcal{S}_i^{(v)}\right), & \text{if } b_i = \mathrm{section}.
\end{cases}
\label{eq:mixed_views}
\end{equation}

In the dropout branch, the two views share the same title--abstract text; in the section branch, the title--abstract view is paired with a genuine cross-section view of the same patent. Because the section type is sampled at the second level rather than in Eq.~\eqref{eq:strategy_sampling}, the dropout share is fixed at one half regardless of how many section types fill \(\mathcal{S}^{(v)}\): a variant with $K$ section types trains with one half dropout pairs and $\tfrac{1}{2K}$ pairs of each section, which keeps the amount of local-consistency signal comparable across single-section and multi-section variants. If $\mathcal{S}_i^{(v)}$ is empty for patent $i$, or the sampled section text is too short after cleaning, the instance falls back to the dropout branch, so the realized dropout share is at least this nominal one half.

Since the strategy and section type are sampled per training instance, a mini-batch contains a mixture of dropout and section pairs, and both view types are optimized under the same in-batch contrastive objective rather than in separate training stages.

The same construction generalizes beyond patent sections: the first level may sample from any variant-specific strategy set \(\mathcal{B}^{(v)}\) containing the dropout strategy, with a resulting dropout share of \(1/|\mathcal{B}^{(v)}|\). Our main method corresponds to \(\mathcal{B}^{(v)} = \{\mathrm{dropout},\ \mathrm{section}\}\); the baseline variants of Section~\ref{sec:positive_variants} instead pair dropout with generic text perturbations or metadata-based matching under the same scheme.

\subsection{Contrastive Objective}
\label{sec:objective}
Let \(g_\theta(\cdot\,;\xi)\) denote the Transformer encoder under dropout mask \(\xi\), and let \(\mathrm{CLS}(\cdot)\) extract the final-layer hidden state of the \texttt{[CLS]} token. During training, every encoding draws an independent dropout mask with rate $p$; dropout is therefore a property of the encoder applied to \emph{all} views, in both the dropout and section branches, rather than an operation attached to a particular view. Following SimCSE~\cite{gao2021simcse}, we use an MLP projection head during contrastive training but discard it for inference.

Specifically, for a mini-batch of $B$ patents, we encode the two views of each patent with independent masks $\xi_i \neq \xi_i'$,
\begin{equation}
c_i^{(1)} = \mathrm{CLS}\!\left(g_\theta(x_i^{(1)};\,\xi_i)\right),
\qquad
c_i^{(2)} = \mathrm{CLS}\!\left(g_\theta(x_i^{(2)};\,\xi_i')\right),
\end{equation}
and compute the training representations
\begin{equation}
z_i^{(1)} = \mathrm{MLP}\!\left(c_i^{(1)}\right),
\qquad
z_i^{(2)} = \mathrm{MLP}\!\left(c_i^{(2)}\right).
\end{equation}
Both representations are $\ell_2$-normalized before computing similarities. In the dropout branch, $x_i^{(1)}$ and $x_i^{(2)}$ are the same text, so the positive pair differs only through the independent masks; in the section branch, the views differ both textually and through the masks.

We train the model using an in-batch contrastive loss:
\begin{equation}
\mathcal{L}_i
=
-\log
\frac{
\exp\!\big(\mathrm{sim}(z_i^{(1)}, z_i^{(2)}) / \tau\big)
}{
\sum_{j=1}^{B}
\exp\!\big(\mathrm{sim}(z_i^{(1)}, z_j^{(2)}) / \tau\big)
},
\label{eq:loss}
\end{equation}
where $\mathrm{sim}(\cdot,\cdot)$ is cosine similarity and $\tau$ is the contrastive temperature.
Views are paired within the same patent, while second-position views from other patents in the mini-batch act as negatives.
The full objective is the average over the mini-batch:
\begin{equation}
\mathcal{L} = \frac{1}{B}\sum_{i=1}^{B}\mathcal{L}_i.
\end{equation}

Because the in-batch negatives in Eq.~\eqref{eq:loss} are drawn exclusively from the second-position views, the two views of each pair are swapped with probability $0.5$ before encoding. For the dropout branch this is a no-op, since both views share the same text; for the section branch, the section view occupies the first position in half of the instances. This symmetrization exposes both title--abstract and section representations as in-batch negatives, rather than restricting the negative pool to a single view type, and is applied uniformly across all strategies, including the baseline perturbation and metadata constructions.

At inference time, dropout is disabled and the MLP projection head is discarded; patent embeddings are obtained from the $\ell_2$-normalized \texttt{[CLS]} representation of the encoder.

\section{Experiments}
\label{sec:experiments}

\subsection{Training Corpus and Section Extraction}
\label{sec:training_data}

We train on the Harvard USPTO Patent Dataset (HUPD)~\cite{suzgun2022hupd}, which contains English-language USPTO utility patent applications filed between 2004 and 2018. Our training split uses applications filed from 2010 to 2018, yielding 2.78M patents after filtering. For each patent, we use the title and extract six textual sections when available: \textit{abstract}, \textit{claims}, \textit{summary}, \textit{background}, \textit{brief description of drawings}, and \textit{detailed description}. For simplicity, we will henceforth use \textit{drawing} to refer to the brief description of drawings, and \textit{description} to refer to the detailed description.

The title is concatenated with the abstract to form the title--abstract view. Patents without a valid title--abstract view are removed. Optional sections are retained when present and discarded when missing; sections shorter than 15 words are treated as missing, except for titles, and section-branch instances whose section pool is empty fall back to the dropout branch at training time. We extract drawing and description sections from the full-description field from HUPD dataset\footnote{\url{https://huggingface.co/datasets/HUPD/hupd}} using heading-based rules. We remove canceled claims, strip source-markup headings and delimiter text from summary and background fields, and discard misclassified summary entries that begin with drawing-description headings such as ``Brief Description of the Drawings.''

Each retained section is prefixed with a section-specific token. We use the special tokens from our initial BERT-for-Patents~\cite{SrebrovicYonamine2020} checkpoint: \texttt{[abstract]}, \texttt{[claim]}, \texttt{[summary]}, and \texttt{[invention]} for abstract, claims, summary, and background, respectively. We additionally introduce \texttt{[drawing]} and \texttt{[description]} for the drawing and description sections, initialized from the \texttt{[invention]} embedding.

\subsection{Training Setup}
\label{sec:training_setup}

Our encoder is initialized from BERT-for-Patents\footnote{\url{https://huggingface.co/anferico/bert-for-patents}} and fine-tuned with a maximum input length of 512 tokens. Contrastive training uses the final-layer \texttt{[CLS]} representation with an MLP projection head; the head is discarded at inference time, and evaluation embeddings are $\ell_2$-normalized before similarity computation. Models are trained for one epoch with batch size 512 on a single NVIDIA H100 GPU. We use AdamW with learning rate $1\times10^{-5}$, cosine decay, 10\% warm-up, fp16 mixed precision, and DeepSpeed Stage~1.

We treat the contrastive temperature $\tau$ as a calibration hyperparameter. The dropout rate $p$ sets the encoder's hidden and attention dropout probabilities and therefore applies to every training-time encoding, in all positive-construction variants. For dropout-only training we additionally sweep $p$, since in that setting the positive pair consists of the same title--abstract text encoded twice, so $p$ directly controls the only source of variation between the two views; note, however, that changing $p$ also changes the regularization strength of the encoder itself. Unless otherwise specified, we use $\tau=0.05$ and $p=0.1$.

\subsection{Positive-Construction Variants}
\label{sec:positive_variants}

We compare several positive-construction strategies under the same in-batch contrastive framework. All mixed variants follow the shared sampling procedure of Section~\ref{sec:mixed_construction} and differ only in their strategy set \(\mathcal{B}^{(v)}\) and, for section variants, the section pool \(\mathcal{S}^{(v)}\). In all self-supervised variants, one view of each positive pair is the title--abstract view of the patent; following Section~\ref{sec:objective}, the two views are randomly swapped with probability $0.5$ before encoding, so both view types appear on both sides of the objective. Views are paired within the same patent or according to the specified construction, while second-position views from other patents in the same mini-batch serve as negatives.

\paragraph{Dropout-only.}
The dropout-only baseline follows the SimCSE-style construction: the positive view is a second stochastic encoding of the same title--abstract input under an independent dropout mask. We vary both the contrastive temperature $\tau$ and the dropout rate $p$, and report the default setting together with representative calibrated configurations in the main results.

\paragraph{Generic text perturbations.}
We evaluate three title--abstract perturbations: sentence shuffling~\cite{yan2021consert}, random cropping, and paraphrasing. Sentence shuffling reorders sentences in the title--abstract view. Random cropping removes a continuous 10\% span. Paraphrasing rewrites the title--abstract view using Qwen3-0.6B~\cite{qwen3technicalreport}. The Crop and Paraphrase variants use \(\mathcal{B} = \{\mathrm{dropout},\ \mathrm{crop}\}\) and \(\{\mathrm{dropout},\ \mathrm{paraphrase}\}\) respectively (one half dropout pairs, one half perturbation pairs), while Crop/shuffle uses \(\mathcal{B} = \{\mathrm{dropout},\ \mathrm{crop},\ \mathrm{shuffle}\}\), so each strategy receives one third of the training instances. Crop and Paraphrase are therefore directly comparable to our single-section variants: both sides train with one half dropout pairs, so the comparison isolates the choice of the second view construction. Crop/shuffle does not support the same controlled comparison, since it differs from the section variants both in its view constructions and in its lower dropout share.

\paragraph{Mixed dropout--section positives.}
Our main construction instantiates the sampling of Section~\ref{sec:mixed_construction} with \(\mathcal{B} = \{\mathrm{dropout},\ \mathrm{section}\}\). Single-section variants (``Ours + claim'' etc.) train with one half dropout pairs and one half pairs of the named section; two-section variants split the section half evenly, e.g., ``Ours + claim/summary'' trains with one half dropout, one quarter claim, and one quarter summary pairs.

\paragraph{IPC-matched positives.}
As a metadata-supervised diagnostic, we also train a variant in which the augmentation strategy pairs the patent with another patent sharing the same IPC subgroup list, i.e., \(\mathcal{B} = \{\mathrm{dropout},\ \mathrm{IPC\text{-}match}\}\). Half of its training pairs are therefore dropout pairs and half link the title--abstract views of two IPC-matched patents; instances without an IPC-matched partner fall back to the dropout branch. This variant is not self-supervised and is not part of our main method, but it helps contextualize how much technology-label supervision changes the embedding geometry and IPC classification behavior.

Unless otherwise noted, ``Ours'' refers to the mixed dropout--section construction. We also report individual and combined section variants to test which patent sections provide the most useful positive views.

\subsection{Baseline Models}
\label{sec:baselines}

We compare against lexical, domain-specific, citation-trained, and general-purpose embedding baselines. BM25 serves as a non-neural lexical retrieval baseline. For neural baselines, we evaluate PatentBERT~\cite{lee2020patent}, BERT-for-Patents~\cite{SrebrovicYonamine2020}, SPECTER~2.0~\cite{singh2022scirepeval}, Pat-SPECTER and PaECTER~\cite{ghosh2024paecter}, and gte-Qwen2-7B-instruct~\cite{li2023towards}. These models cover different sources of supervision and domain adaptation: patent-domain pretraining, CPC or citation supervision, transfer from scientific document embeddings, and large-scale general-purpose embedding training. BERT-for-Patents is also used as the initialization checkpoint for our contrastive models.

For fair comparison, neural embedding baselines are evaluated using the same patent views as our models whenever possible. In title--abstract settings, we encode the concatenated title and abstract. In section-aware retrieval settings, we encode each required patent section independently with the same model and aggregate section similarities at evaluation time.
BM25 uses the same query and candidate text fields as the corresponding retrieval setting.

\section{Evaluation and Results}
\label{sec:evaluation}

We evaluate frozen patent representations on prior-art retrieval and IPC subclass classification, without task-specific fine-tuning on any evaluation benchmark. Retrieval measures fine-grained relevance between query patents and candidate prior art, while IPC classification measures broader technology-domain organization. Together, they test whether the learned embeddings support both prior-art ranking and technology analysis.

\subsection{Evaluation Protocol}
\label{sec:evaluation_protocol}

\paragraph{EPO search-report retrieval.}
We construct a prior-art retrieval benchmark from 200 EPO applications filed in 2021--2022. Relevant documents are extracted from official EPO search reports, which label cited documents by examiner relevance category. Following the PatentMatch extraction setting~\cite{risch2020patentmatch}, we retain citations labeled $X$, $Y$, or $A$, where $X$ denotes novelty-destroying prior art, $Y$ denotes documents relevant in combination, and $A$ denotes technological background.

To avoid an overly easy retrieval pool based on random negatives, we combine examiner-cited documents with hard negatives from three sources. First, \emph{more-like-this} negatives are retrieved with Elasticsearch using the query patent's title, abstract, and citing claim against title, abstract, claims, and description fields. These documents are lexically close to the query but are not examiner-cited. Second, \emph{IPC-matching} negatives are patents from the same IPC subclass as the query that are not cited, testing fine-grained discrimination within the same technology area. Third, \emph{cited-of-cited} negatives are patents cited by the query's cited documents but not directly cited by the query, creating citation-neighborhood competitors. Negatives are pooled globally into a shared retrieval pool, rather than forming small per-query reranking sets. The final pool contains 47,636 candidate documents, with an average of 2.5 examiner-cited positives per query and hard negatives from the three sources.

We evaluate two query--candidate configurations. In \textbf{Abs$\rightarrow$Abs}, both queries and candidates are represented by title-plus-abstract embeddings, reflecting summary-level novelty search. In \textbf{Clm$\rightarrow$All}, queries are represented by claims, while each candidate patent is represented by title-plus-abstract, claims, and description embeddings. We score each candidate patent by max aggregation over its three section embeddings. This setting evaluates claim-to-disclosure matching and is closer to patentability search and examination workflows, where claims are compared against different parts of potential prior-art documents.

Because EPO search reports distinguish levels of examiner relevance, we report graded nDCG using linear gains $X=3$, $Y=2$, and $A=1$. For a query $q$, let $g_{q,r}$ be the gain of the document ranked at position $r$. We compute
\begin{equation}
\mathrm{DCG@}K(q)=\sum_{r=1}^{K}\frac{g_{q,r}}{\log_2(r+1)},
\quad
\mathrm{nDCG@}K(q)=
\frac{\mathrm{DCG@}K(q)}{\mathrm{IDCG@}K(q)},
\end{equation}
where $\mathrm{IDCG@}K(q)$ is the maximum possible DCG@K obtained by sorting the relevant documents for $q$ by gain. If a query--candidate pair has multiple citation categories, we keep the maximum gain. To remain comparable with binary retrieval evaluations, we also report Recall@100 by treating all examiner-cited documents as relevant. We additionally compute untruncated mAP and release it with the evaluation outputs. All retrieval metrics are macro-averaged over queries.

\paragraph{DAPFAM family-level retrieval.}
We further evaluate on DAPFAM~\cite{ayaou2025dapfam}, a public domain-aware patent retrieval benchmark with 1,247 query families and 45,336 target families. Compared with our EPO search-report benchmark, DAPFAM uses patent families rather than individual publications as retrieval units, binary citation-based relevance rather than graded examiner categories, and in-domain/out-of-domain partitions based on IPC3 overlap. Following the DAPFAM document-level setting, we encode each query and candidate family using title, abstract, and claims (TAC), and apply the same TAC input to all dense baselines and our models. We report nDCG@100 and Recall@100 following the DAPFAM protocol. No DAPFAM relevance labels or family links are used during training. We also run a title--abstract-only variant as a control aligned with our EPO Abs$\rightarrow$Abs setting and release those results with the evaluation outputs.

\paragraph{IPC subclass classification.}
To evaluate whether the learned embeddings preserve technology-domain structure, we perform IPC subclass classification with a $k$-nearest-neighbor classifier using $k=10$. We use 30k USPTO patents from HUPD 2005--2009, with an 85/15 train/test split and 555 unique IPC subclasses in total. Each patent is represented by its frozen title--abstract embedding, and the predicted label is assigned by majority vote among its nearest neighbors. We report Precision@1, which is equivalent to top-1 accuracy in this single-label setting. We use KNN rather than a learned linear probe because our goal is to evaluate the geometry of the same frozen embedding space used in the retrieval experiments, without introducing an additional supervised classifier.

\subsection{Main EPO Prior-art Search Results}
\label{sec:main_results}

\begin{table*}[h]
\centering
\small
\setlength{\tabcolsep}{3.4pt}
\caption{
Main results on EPO search-report retrieval and IPC subclass classification.
nDCG uses graded examiner relevance with linear gains $X=3$, $Y=2$, and $A=1$.
R@100 and mAP treat all examiner-cited documents as binary relevant.
IPC P@1 is KNN classification precision using title--abstract embeddings.
Bold and underline indicate the best and second-best results within this table.
}
\label{tab:main_results}
\begin{tabular}{lccccccc}
\toprule
\multirow{2}{*}{Model}
& \multicolumn{3}{c}{\textbf{Abs$\rightarrow$Abs}}
& \multicolumn{3}{c}{\textbf{Clm$\rightarrow$All}}
& \textbf{IPC} \\
\cmidrule(lr){2-4}\cmidrule(lr){5-7}\cmidrule(lr){8-8}
& nDCG@10 & R@100 & mAP
& nDCG@10 & R@100 & mAP
& P@1 \\
\midrule
\multicolumn{8}{l}{\textit{Retrieval and pretrained-encoder baselines}} \\
BM25 & 0.1476 & 0.5726 & 0.1149 & 0.1198 & 0.5564 & 0.0874 & -- \\
SPECTER~2.0 & 0.1452 & 0.4976 & 0.1063 & 0.0878 & 0.4935 & 0.0691 & 52.54 \\
BERT-for-Patents & 0.1307 & 0.5604 & 0.0972 & 0.1046 & 0.4909 & 0.0783 & 57.09 \\
Pat-SPECTER & 0.1684 & 0.6629 & 0.1246 & 0.1233 & 0.6534 & 0.0965 & 55.40 \\
PaECTER & \underline{0.2086} & \underline{0.7594} & \underline{0.1612} & \textbf{0.1822} & \textbf{0.7694} & \textbf{0.1370} & 60.47 \\
gte-Qwen2-7B-instruct & 0.1898 & 0.6902 & 0.1420 & \underline{0.1770} & 0.7349 & \underline{0.1365} & 60.99 \\

\midrule
\multicolumn{8}{l}{\textit{Dropout-only contrastive baselines}} \\
Dropout ($\tau$=0.05, drop=0.1)   & 0.1547 & 0.6045 & 0.1180 & 0.0753 & 0.3970 & 0.0566 & 48.42 \\
Dropout ($\tau$=0.0001, drop=0.3) & 0.1963 & 0.7514 & 0.1465 & 0.0825 & 0.4651 & 0.0646 & 61.55 \\
Dropout ($\tau$=0.1, drop=0.1)    & 0.1324 & 0.5781 & 0.0998 & 0.1234 & 0.5432 & 0.0927 & 51.60 \\
Dropout ($\tau$=0.5, drop=0.1)    & 0.1092 & 0.5489 & 0.0860 & 0.1048 & 0.5313 & 0.0834 & 55.18 \\
\midrule
\multicolumn{8}{l}{\textit{Metadata-supervised diagnostic}} \\
IPC-subgroup-match positives & 0.1453 & 0.6131 & 0.1131 & 0.1166 & 0.6573 & 0.0905 & \textbf{63.19} \\

\midrule
\multicolumn{8}{l}{\textit{Generic title--abstract augmentations}} \\
Crop & 0.1975 & 0.6944 & 0.1522 & 0.1344 & 0.6151 & 0.1022 & 56.44 \\
Crop/shuffle & 0.1975 & 0.7036 & 0.1543 & 0.1384 & 0.6455 & 0.1074 & 56.98 \\
Paraphrase & 0.1855 & 0.7137 & 0.1455 & 0.1404 & 0.6305 & 0.1091 & 54.64 \\

\midrule
\multicolumn{8}{l}{\textit{Single patent-section positives}} \\
Ours + claim & \textbf{0.2189} & \textbf{0.7619} & \textbf{0.1636} & 0.1753 & \underline{0.7470} & 0.1331 & 60.96 \\
Ours + summary & 0.2039 & 0.7455 & 0.1543 & 0.1682 & 0.7262 & 0.1249 & 61.44 \\
Ours + background & 0.1935 & 0.7082 & 0.1380 & 0.1633 & 0.6991 & 0.1208 & 62.29 \\
Ours + drawings & 0.1987 & 0.6791 & 0.1424 & 0.1680 & 0.7245 & 0.1258 & 60.65 \\
Ours + description & 0.1862 & 0.6730 & 0.1393 & 0.1617 & 0.6989 & 0.1210 & 61.86 \\

\midrule
\multicolumn{8}{l}{\textit{Two-section positive pools}} \\
Ours + claim/summary & 0.2082 & 0.7532 & 0.1594 & 0.1698 & 0.7356 & 0.1295 & 60.96 \\
Ours + claim/background & 0.2006 & 0.7297 & 0.1442 & 0.1680 & 0.7307 & 0.1258 & 62.63 \\
Ours + claim/drawings & 0.2066 & 0.7026 & 0.1528 & 0.1672 & 0.7357 & 0.1272 & 61.01 \\
Ours + claim/description & 0.2010 & 0.6984 & 0.1472 & 0.1643 & 0.7212 & 0.1258 & 61.50 \\
Ours + summary/background & 0.2003 & 0.7332 & 0.1422 & 0.1658 & 0.7166 & 0.1234 & \underline{62.74} \\
Ours + summary/drawings & 0.2029 & 0.6987 & 0.1475 & 0.1637 & 0.7184 & 0.1222 & 61.03 \\
Ours + summary/description & 0.1957 & 0.6791 & 0.1437 & 0.1647 & 0.7103 & 0.1226 & 62.16 \\
Ours + background/description & 0.1920 & 0.6784 & 0.1417 & 0.1645 & 0.6988 & 0.1231 & 62.34 \\
Ours + background/drawings & 0.2021 & 0.7028 & 0.1426 & 0.1690 & 0.7242 & 0.1237 & 62.61 \\
Ours + drawings/description & 0.1935 & 0.6762 & 0.1418 & 0.1665 & 0.7066 & 0.1281 & 61.32 \\
\bottomrule
\end{tabular}
\end{table*}

Table~\ref{tab:main_results} compares lexical retrieval, pretrained patent encoders, calibrated dropout-only contrastive models, generic text perturbations, metadata-matched positives, and our section-based positives on EPO search-report retrieval and IPC subclass classification. The IPC-subgroup-match row uses exact IPC subgroup-list matches to construct positives; it is included only as a metadata-supervised diagnostic and is not part of our self-supervised method.

First, dropout-only contrastive learning becomes substantially stronger after calibration, but its best configuration is evaluation-dependent. For Abs$\rightarrow$Abs retrieval and IPC classification, low temperatures combined with a high dropout rate are strongly preferred: the $\tau$=0.0001, drop=0.3 configuration reaches 0.1963 nDCG@10 and 61.55 IPC P@1, although this configuration lies at the boundary of our sweep, and most other low-temperature settings plateau around 0.178--0.183. Both remain weak in Clm$\rightarrow$All retrieval. Conversely, Clm$\rightarrow$All retrieval favors much higher temperatures, with the best dropout-only result (0.1234 nDCG@10) at $\tau$=0.1---a setting that substantially reduces Abs$\rightarrow$Abs and IPC performance. This indicates that the best dropout-only calibration is tied to the evaluation setting: configurations that improve title--abstract retrieval do not transfer to claim-to-disclosure retrieval, and vice versa.

Second, generic title--abstract augmentations improve over the default dropout-only model, especially in Clm$\rightarrow$All retrieval, showing that non-identical positives are useful. The Crop and Paraphrase baselines mix their perturbation with dropout in the same one-half/one-half proportion as our single-section variants, so those comparisons hold the local-consistency component fixed and isolate the choice of the second view construction; Crop/shuffle uses a lower dropout share of one third. Under these comparisons, generic perturbations remain below the strongest section-positive models, suggesting that patent-internal sections provide more effective view variation than perturbations of the title--abstract view alone.

Third, the IPC-subgroup-match diagnostic shows that IPC-supervised positives emphasize a different notion of relatedness from examiner prior-art relevance. This variant achieves the best IPC P@1 in the table, as expected because it uses exact IPC subgroup-list matches during training. However, it does not improve Abs$\rightarrow$Abs retrieval over the calibrated dropout baselines or generic perturbations, and remains below section-based positives. It achieves comparatively strong Clm$\rightarrow$All R@100 (0.6573), above the dropout-only and generic-perturbation baselines, but its Clm$\rightarrow$All nDCG@10 remains below them. This suggests that IPC subgroup supervision is effective for technology-domain clustering, but does not consistently provide the same fine-grained prior-art retrieval signal as claim-based section positives.

Fourth, patent-section positives give the strongest Abs$\rightarrow$Abs retrieval results and competitive Clm$\rightarrow$All performance. The claim-positive model achieves the best Abs$\rightarrow$Abs nDCG@10, R@100, and mAP in the table. In Clm$\rightarrow$All retrieval, PaECTER remains the strongest model, but claim-based section positives are close to the general-purpose gte-Qwen2-7B-instruct embedding model and clearly improve over dropout-only and generic perturbation baselines. This comparison is important because our section-based models do not use IPC labels, citation links, or external relevance annotations during training.

Finally, retrieval and IPC classification favor different patent sections. Claim positives are most effective for prior-art retrieval, consistent with the role of claims in defining the protected invention. In contrast, section pools involving background achieve the highest IPC P@1: summary/background gives the best IPC result, and claim/background gives the second-best result. This pattern suggests that claims provide fine-grained invention-level signals for retrieval, while background sections contribute broader technology-domain context that is useful for subclass-level organization.

\subsection{Dropout and Section-Pair Positives Calibrate Differently}
\label{sec:temperature}

\begin{figure}[ht]
\centering
\includegraphics[width=\linewidth]{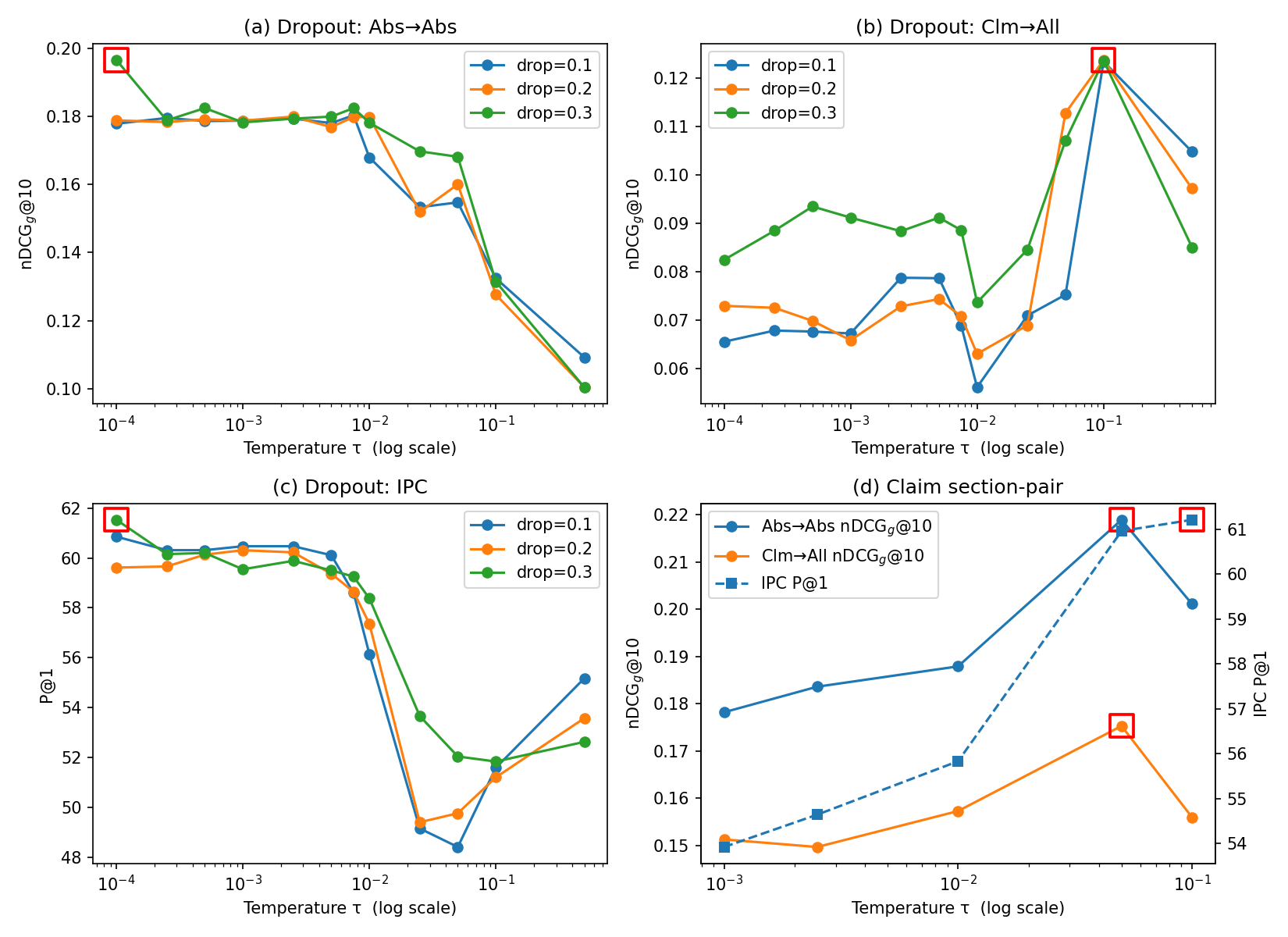}
\caption{
Calibration of dropout-only and claim section-pair contrastive training. 
Panels (a)--(c) show dropout-only models across temperatures and dropout rates. Abs$\rightarrow$Abs retrieval plateaus at low temperatures and declines beyond $\tau \approx 0.01$; Clm$\rightarrow$All retrieval instead improves with temperature and peaks at $\tau$=0.1; IPC classification is strongest on the low-temperature plateau, collapses at intermediate temperatures, and partially recovers at very high temperature. Red boxes mark the best configuration in each panel. Panel (d) shows the claim section-pair model, for which both retrieval and IPC performance improve with temperature, peaking at $\tau$=0.05--0.1.
}
\label{fig:calibration_sweep}
\end{figure}

Table~\ref{tab:main_results} shows that calibrated dropout-only training can be competitive, but its best setting differs across Abs$\rightarrow$Abs retrieval, Clm$\rightarrow$All retrieval, and IPC classification. Because the main table reports only representative configurations, we examine the broader calibration trends in Figure~\ref{fig:calibration_sweep}. We vary temperature and dropout rate for dropout-only training, and compare these trends with claim section-pair training as a representative section-based configuration.

Figure~\ref{fig:calibration_sweep} shows that dropout-only training is highly sensitive to calibration. For Abs$\rightarrow$Abs retrieval, most low-temperature configurations ($\tau \le 0.0075$) plateau at a similar level, with the strongest result at the boundary of our sweep ($\tau$=0.0001, drop=0.3); performance degrades steadily for $\tau \ge 0.01$. IPC classification shares the low-temperature preference but is not monotonic in temperature: it collapses at intermediate temperatures ($\tau$=0.025--0.05) and partially recovers at $\tau$=0.5, without regaining the low-temperature level. However, these configurations do not transfer to Clm$\rightarrow$All retrieval, where performance instead increases with temperature and peaks at $\tau$=0.1.
This pattern suggests that a very sharp dropout-only contrastive objective strengthens local invariance within the title--abstract view, which benefits Abs$\rightarrow$Abs retrieval and, on the low-temperature plateau, IPC classification, but makes the learned geometry less suitable for comparing claims against different disclosure sections.

Claim section-pair training follows a different calibration pattern. In the available sweep, claim-positive models perform best at higher temperatures, with the strongest retrieval results around $\tau=0.05$ and competitive IPC performance at $\tau=0.1$. Lower temperatures such as $\tau=0.01$, $0.0025$, and $0.001$ underperform across Abs$\rightarrow$Abs retrieval, Clm$\rightarrow$All retrieval, and IPC classification. This contrasts with dropout-only training, where low-temperature configurations are strongest for Abs$\rightarrow$Abs retrieval and IPC classification. The difference is expected because the two positive constructions define different alignment problems. Dropout positives are two stochastic encodings of the same title--abstract text, so a sharper contrastive distribution can help enforce local invariance. Claim positives, in contrast, connect the title--abstract view to a legally and lexically different section of the same patent. These positives should be aligned at the invention level, but they are not near-duplicates. A very low temperature may therefore make the alignment constraint too strict for cross-section positives, reducing the model's ability to preserve useful section-specific information.

Taken together, these sweeps show that calibration depends on the positive view being optimized. Low-temperature dropout training improves the title--abstract space, but does not carry over to claim-to-disclosure retrieval. Claim section-pair training works best at a higher temperature than the tuned dropout-only baseline, suggesting that cross-section positives should be aligned as related views of the same invention rather than compressed into near-duplicate representations.

\subsection{Retrieval Performance on DAPFAM}
\label{sec:dapfam_results}

\begin{table}[t]
\centering
\small
\setlength{\tabcolsep}{4pt}
\caption{
External retrieval on DAPFAM under the benchmark title--abstract--claims (TAC) input setting.
DAPFAM evaluates retrieval at the patent-family level.
We report the overall and out-of-domain partitions; full partition-level outputs, including title--abstract-only controls, are released with the code.
Bold and underline indicate the best and second-best results.
}
\label{tab:dapfam}
\begin{tabular}{l|cc|cc}
\toprule
\multirow{2}{*}{Model}
& \multicolumn{2}{c|}{\textbf{All}}
& \multicolumn{2}{c}{\textbf{Out-of-domain}} \\
& nDCG@100 & R@100
& nDCG@100 & R@100 \\
\midrule
BM25                         & 0.2287 & 0.2638 & 0.0384 & 0.0961 \\
SPECTER~2.0                  & 0.2389 & 0.2875 & 0.0330 & 0.0860 \\
BERT-for-Patents             & 0.2246 & 0.2702 & 0.0366 & 0.0973 \\
Pat-SPECTER                  & 0.2692 & 0.3242 & 0.0421 & 0.1151 \\
PaECTER                      & 0.3361 & 0.4067 & 0.0568 & 0.1595 \\
gte-Qwen2-7B-instruct         & 0.3430 & 0.4110 & 0.0586 & 0.1485 \\
\midrule
Dropout ($\tau$=0.05, drop=0.1)   & 0.1495 & 0.1682 & 0.0357 & 0.0762 \\
Dropout ($\tau$=0.0001, drop=0.3) & 0.3253 & 0.3929 & 0.0561 & 0.1527 \\
Dropout ($\tau$=0.1, drop=0.1)    & 0.2209 & 0.2558 & 0.0436 & 0.1108 \\
Dropout ($\tau$=0.5, drop=0.1)    & 0.2482 & 0.3079 & 0.0427 & 0.1119 \\
\midrule
Crop/shuffle           & 0.2617 & 0.3032 & 0.0470 & 0.1153 \\
\midrule
Ours + claim                  & \textbf{0.3495} & \textbf{0.4195} & \textbf{0.0647} & \textbf{0.1722} \\
Ours + summary                & 0.3451 & 0.4132 & 0.0620 & 0.1661 \\
Ours + claim/summary          & \underline{0.3471} & \underline{0.4168} & \underline{0.0637} & \underline{0.1706} \\
\bottomrule
\end{tabular}
\end{table}

Table~\ref{tab:dapfam} evaluates the models on DAPFAM using the benchmark title--abstract--claims input setting. DAPFAM complements our EPO benchmark through its family-level and domain-aware design: it ranks patent families rather than individual publications, and explicitly separates in-domain from out-of-domain retrieval according to IPC3 overlap. The out-of-domain partition is particularly relevant here because it tests whether models can retrieve cited patent families even when query and target families do not share the same IPC3 domain.

Section-based positives perform best on both the overall and out-of-domain partitions. Ours + claim achieves the highest nDCG@100 and Recall@100, outperforming PaECTER and gte-Qwen2-7B-instruct under the same TAC input format, and the same ordering holds out of domain. Calibrated dropout-only training is also strong on DAPFAM: the $\tau$=0.0001, drop=0.3 configuration approaches PaECTER on the overall partition. Section positives nevertheless remain ahead on every partition, with the largest relative margin out of domain, indicating that patent-internal sections provide a training signal for family-level retrieval beyond what dropout calibration alone achieves.

\subsection{Cross-Section Alignment Diagnostics}
\label{sec:section_alignment}

The calibration results in Section~\ref{sec:temperature} suggest that dropout-only training mainly improves the title--abstract space, whereas section-pair training directly exposes the encoder to different sections of the same invention. To examine this effect, we measure cross-section compatibility on the 47,836 patent documents used in our EPO retrieval benchmark (the 47,636-document candidate pool plus the 200 query patents). This provides a diagnostic view of section-aware retrieval, where different patent sections may be compared at inference time.

For a patent $d$, let $S_d$ denote its available sections and let $z_{d,s}$ be the $\ell_2$-normalized embedding of section $s\in S_d$. We define the intra-document section distance as the average cosine distance over unordered section pairs from the same patent:
\begin{equation}
A_{\mathrm{intra}}(d)
=
\mathbb{E}_{(s_i,s_j)\in \mathcal{P}(S_d)}
\left[
1-\cos(z_{d,s_i},z_{d,s_j})
\right],
\end{equation}
where $\mathcal{P}(S_d)$ is the set of unordered pairs of distinct available sections in $S_d$.

Because models differ in their global embedding spread, we normalize this intra-document distance by the expected cosine distance between randomly sampled sections from different patents:
\begin{equation}
\mathrm{IDA\text{-}Ratio}
=
\frac{\mathbb{E}_{d}[A_{\mathrm{intra}}(d)]}
{\mathbb{E}_{(i,j)\sim P_{\mathrm{rand}}}
\left[
1-\cos(z_i,z_j)
\right]},
\end{equation}
where $P_{\mathrm{rand}}$ samples section embeddings from different patents. Lower values indicate that sections of the same patent are closer to one another relative to the model's overall embedding spread.

\begin{table}[ht]
\centering
\small
\setlength{\tabcolsep}{4pt}
\definecolor{idaA}{gray}{0.82} 
\definecolor{idaB}{gray}{0.88} 
\definecolor{idaC}{gray}{0.93} 
\definecolor{idaD}{gray}{0.97} 
\caption{
Intra-document alignment ratio (IDA-Ratio) for representative models.
For each section pair, we report the mean cosine distance between two sections of the same patent, normalized by the mean cosine distance of random cross-document section pairs from the same model.
Values below $1$ indicate stronger within-patent alignment than random cross-document pairs; lower is better.
Cell shading is applied to all numeric cells; darker cells indicate lower IDA-Ratio values.
}
\label{tab:ida}
\begin{tabular}{lcccc}
\toprule
Model & Abs--Clm & Abs--Desc & Clm--Desc & Avg. \\
\midrule
\multicolumn{5}{l}{\textit{Pretrained and retrieval baselines}} \\
BERT-for-Patents
& \cellcolor{idaD}0.9150
& \cellcolor{idaD}0.8782
& \cellcolor{idaA}0.2748
& \cellcolor{idaD}0.6893 \\
PaECTER
& \cellcolor{idaB}0.3217
& \cellcolor{idaB}0.4090
& \cellcolor{idaA}0.2483
& \cellcolor{idaB}0.3263 \\
gte-Qwen2-7B-instruct
& \cellcolor{idaB}0.3487
& \cellcolor{idaB}0.3926
& \cellcolor{idaB}0.3139
& \cellcolor{idaB}0.3517 \\
\midrule
\multicolumn{5}{l}{\textit{Dropout-only contrastive baselines}} \\
Dropout ($\tau{=}0.05$, drop$=0.1$)
& \cellcolor{idaD}0.8619
& \cellcolor{idaD}0.7584
& \cellcolor{idaC}0.4964
& \cellcolor{idaD}0.7056 \\
Dropout ($\tau{=}0.0001$, drop$=0.3$)
& \cellcolor{idaD}0.7766
& \cellcolor{idaC}0.5869
& \cellcolor{idaB}0.3744
& \cellcolor{idaC}0.5793 \\
Dropout ($\tau{=}0.1$, drop$=0.1$)
& \cellcolor{idaC}0.5178
& \cellcolor{idaC}0.5917
& \cellcolor{idaC}0.5685
& \cellcolor{idaC}0.5593 \\
Dropout ($\tau{=}0.5$, drop$=0.1$)
& \cellcolor{idaB}0.3520
& \cellcolor{idaB}0.3845
& \cellcolor{idaB}0.3403
& \cellcolor{idaB}0.3589 \\
\midrule
\multicolumn{5}{l}{\textit{Generic title--abstract augmentation}} \\
Crop/shuffle
& \cellcolor{idaC}0.5572
& \cellcolor{idaC}0.5294
& \cellcolor{idaB}0.4253
& \cellcolor{idaC}0.5040 \\
\midrule
\multicolumn{5}{l}{\textit{Patent-section positives}} \\
Ours + claim
& \cellcolor{idaA}0.2741
& \cellcolor{idaB}0.3109
& \cellcolor{idaA}0.2087
& \cellcolor{idaA}0.2646 \\
Ours + summary
& \cellcolor{idaA}0.2790
& \cellcolor{idaB}0.3011
& \cellcolor{idaA}0.2109
& \cellcolor{idaA}0.2637 \\
Ours + background
& \cellcolor{idaA}0.2534
& \cellcolor{idaA}0.2609
& \cellcolor{idaA}\textbf{0.1867}
& \cellcolor{idaA}\textbf{0.2337} \\
Ours + claim/summary
& \cellcolor{idaA}0.2696
& \cellcolor{idaB}0.3059
& \cellcolor{idaA}0.2012
& \cellcolor{idaA}0.2589 \\
Ours + claim/background
& \cellcolor{idaA}0.2569
& \cellcolor{idaA}0.2627
& \cellcolor{idaA}0.1934
& \cellcolor{idaA}0.2377 \\
Ours + summary/background
& \cellcolor{idaA}0.2700
& \cellcolor{idaA}0.2683
& \cellcolor{idaA}0.1954
& \cellcolor{idaA}0.2446 \\
Ours + background/drawings
& \cellcolor{idaA}\textbf{0.2514}
& \cellcolor{idaA}\textbf{0.2595}
& \cellcolor{idaA}0.1978
& \cellcolor{idaA}0.2362 \\
\bottomrule
\end{tabular}
\end{table}

Table~\ref{tab:ida} shows that section-pair models yield the lowest average IDA-Ratios. BERT-for-Patents obtains a relatively low Clm--Desc ratio, but its Abs--Clm and Abs--Desc ratios remain high. This suggests that claims and descriptions may be easier to align because they are both long, technical, and lexically overlapping, whereas aligning abstracts with longer sections requires bridging stronger differences in discourse role, length, and style.

Within dropout-only training, cross-section alignment improves with temperature but never reaches the section-pair level in the regime where retrieval is strong. The retrieval-strongest low-temperature configuration ($\tau$=0.0001, drop=0.3) leaves abstracts poorly aligned with claims (Abs--Clm ratio 0.78), as does the default setting (0.86). Alignment improves at $\tau$=0.1 (0.52) and $\tau$=0.5 (0.35), mirroring the calibration split of Section~\ref{sec:temperature}: $\tau$=0.1 also yields the best dropout-only Clm$\rightarrow$All retrieval---consistent with claim-to-disclosure matching benefiting from cross-section compatibility---but weak Abs$\rightarrow$Abs retrieval, while at $\tau$=0.5, where alignment approaches the section-pair level, all retrieval settings degrade. No single temperature gives dropout-only training both strong alignment and strong retrieval across settings; we leave a mechanistic account of this trade-off to future work.

Section-pair positives consistently bring abstracts, claims, and descriptions of the same invention closer together after accounting for each model's global embedding spread. Background-oriented positives obtain the strongest IDA-Ratios, while claim positives give the strongest retrieval results in Table~\ref{tab:main_results}. This is not contradictory: IDA-Ratio measures broad within-patent coherence, whereas prior-art retrieval also depends on which section carries the most discriminative relevance signal. IDA-Ratio therefore provides a complementary view of the retrieval results in Tables~\ref{tab:main_results} and~\ref{tab:dapfam}: section-pair training improves cross-section compatibility, while retrieval performance still depends on the task-specific relevance signal carried by each section.

\section{Conclusion}
\label{sec:conclusion}

We studied self-supervised patent representation learning through the lens of positive-view construction. A calibrated dropout-only contrastive baseline is substantially stronger than a default SimCSE-style setting, approaching supervised or metadata-informed baselines in some title--abstract retrieval and IPC classification settings. However, its best configuration is evaluation-dependent and does not transfer uniformly to claim-to-disclosure retrieval, where queries and candidate sections differ more strongly in discourse role, length, and lexical style. This shows that, for long and structured patent documents, positive-view construction should be considered together with contrastive temperature and dropout rate.

We introduced mixed dropout--section positives, where each positive pair links the title--abstract view of a patent either to a dropout re-encoding of itself or to another section of the same patent, with the two branches sampled in equal proportion. This uses patent-internal structure as a training-time signal, without IPC labels, citations, or relevance annotations. Across EPO search-report retrieval, DAPFAM family-level retrieval, and IPC KNN classification, section-pair positives improve over dropout-only and generic title--abstract augmentations, and are competitive with strong citation-informed patent encoders and the general-purpose gte-Qwen2-7B-instruct embedding model. The results also show that different sections emphasize different notions of relatedness: claims are most useful for fine-grained prior-art retrieval, while background-oriented positives better capture broader technology-domain structure.

Our diagnostics indicate that these gains are associated with better compatibility among different textual views of the same invention. Section-pair models produce lower intra-document alignment ratios than dropout-only and generic augmentation baselines, indicating stronger alignment between abstracts, claims, and descriptions after accounting for global embedding spread. This complements the retrieval results by showing that section-pair training improves the cross-section geometry most relevant to claim-to-disclosure matching.

\paragraph{Limitations.}
Our findings should be interpreted with several limitations in mind. First, our training data is limited to English USPTO applications. Although we evaluate cross-jurisdictionally on EPO search-report pairs and externally on DAPFAM, we do not yet test multilingual training or broader non-US patent corpora. Extending both training and evaluation across jurisdictions and languages would provide a stronger assessment of robustness and generalization. Second, our method depends on reliable section extraction; noisy, missing, or inconsistently segmented sections may reduce the quality of section positives. Finally, we do not filter patent families during pretraining, so related filings may occasionally appear as negatives in the contrastive pool. This introduces label noise and may lead us to underestimate the model's true potential.



\begin{acks}
We are grateful to the anonymous reviewers for their valuable comments, which helped us refine this paper. We thank Younes Djemmal for constructing the prior-art search evaluation dataset, and also thank him, Kirian Guiller, and Thomas Gerald for their discussions and valuable feedback. This work was partly funded by the last author’s chair in the PRAIRIE institute funded by the French national agency ANR as part of the “Investissements d’avenir” programme under the reference ANR-19-P3IA-0001.
\end{acks}


\bibliographystyle{ACM-Reference-Format}
\balance
\bibliography{sample-base}

\end{document}